\documentclass[10pt,twocolumn,letterpaper]{article}

\usepackage{amssymb} % Equations
\usepackage{amsmath}
\usepackage[percent]{overpic} % Better figures
\usepackage{subcaption}
\usepackage[ruled,linesnumbered]{algorithm2e}

\usepackage{titling}
\usepackage[margin=1in]{geometry}

\DeclareMathOperator*{\argmin}{arg\,min}
\title{Physics-driven Fire Modeling from Multi-view Images}
	
\newcommand{\sourceCodeURL}{https://github.com/Garoe/bath-fire-shader}

\date{}

\begin{document}

	\author{Garoe Dorta \quad Luca Benedetti \quad Dmitry Kit \quad Yong-Liang Yang \\
	% For Computer Graphics Forum: Please use the abbreviation of your first name.
		University of Bath}

\twocolumn[{%
\renewcommand\twocolumn[1][]{#1}%
\maketitle
\begin{center}
   \centering
   \includegraphics[height=4.7cm]{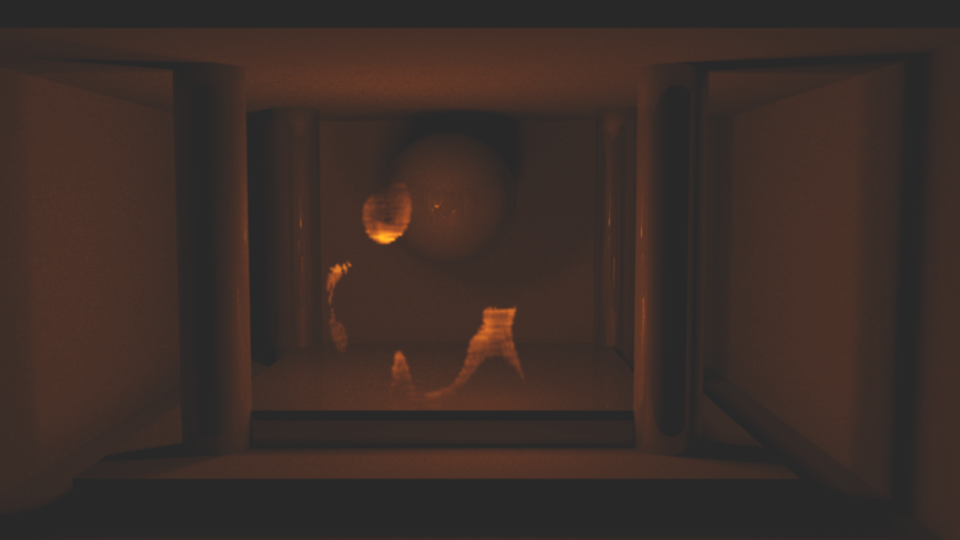}
   \includegraphics[height=4.7cm]{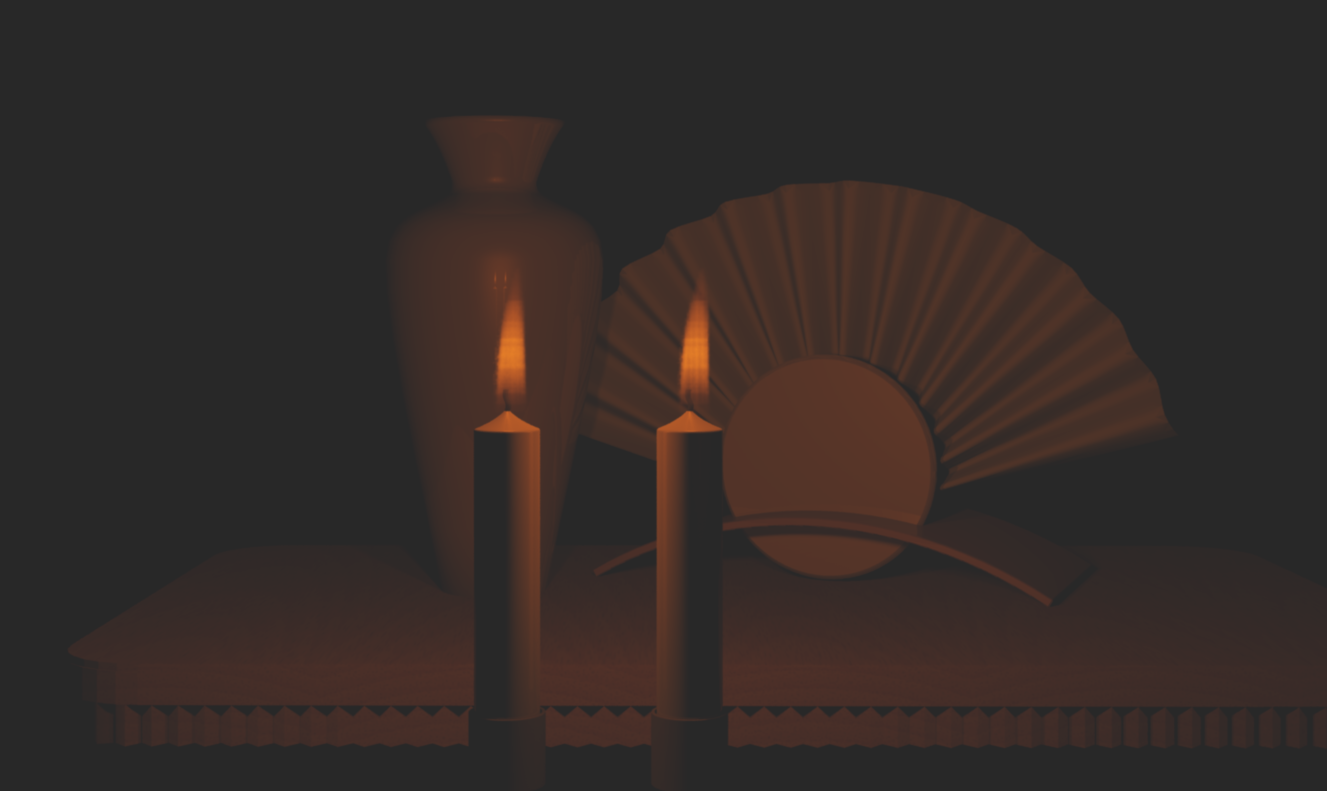}
   \captionof{figure}{Our image-based framework is able to generate realistic flames, which can be placed in a variety of scenes where they provide lifelike illumination. Our algorithm infers physically-based plausible temperature and fuel density values from RGB data input from recorded real fires. These images have been overexposed for display purposes, please refer to Fig.~\ref{fig:teaser-original} to see the unaltered output of our method.}
   \label{fig:teaser}
\end{center}%
}]

\begin{abstract}
Fire effects are widely used in various computer graphics applications such as visual effects and video games. Modeling the shape and appearance of fire phenomenon is challenging as the underlying effects are driven by complex laws of physics.
State-of-the-art fire modeling techniques rely on sophisticated physical simulations which require intensive parameter tuning, or use simplifications which produce physically invalid results.
In this paper, we present a novel method of reconstructing physically valid fire models from multi-view stereo images. Our method, for the first time, provides plausible estimation of physical properties (e.g., temperature, density) of a fire volume using RGB cameras. This allows for a number of novel phenomena such as global fire illumination effects.
The effectiveness and usefulness of our method are tested by generating fire models from a variety of input data, and applying the reconstructed fire models for realistic illumination of virtual scenes.

%Fire effects are widely used in the visual effects and video games industries.
%Controlling the shape and appearance of a flame volume is challenging as the underlying effects are driven by complex chemical reactions in the flames.
%Current techniques rely on sophisticated simulations to generate the data or use simplifications which produce unrealistic effects.
%In this paper we present a new method for automatically estimating physically plausible temperature and density volumetric fields from a pair of stereo images of real flames.
%The advantages of using physically plausible values allow for a number of phenomena such as global illumination effects.
%A novel energy function which incorporates global image terms and local constrains for the volume properties is provided.
%The effectiveness of our method is tested with a variety of input data.

\end{abstract}

% optional quote
%\emph{The fire is the main comfort of the camp, whether in summer or winter, and is about as ample at one season as at another. It is as well for cheerfulness as for warmth and dryness.\\
%Henry David Thoreau}
% Another optional quote
%\emph{A case that can be made for fire being, next to the life process, the most complex of phenomena to understand.\\
% Hoyt Hottle}

\section{Introduction}

Fire is one of the fundamental pillars of our civilization. For more than a million years, we used fire in a variety of applications ranging from the most basic ones like protection, warmth, and food processing, to advanced technological ones~\cite{berna2012microstratigraphic}.
Fire however is not just a tool for us, it shaped our culture~\cite{goudsblom1992fire} and we have a fascination towards its attractive presence and its dangerous nature.
It is thus not surprising that modern digital techniques do not only look into simulation and rendering of fire for engineering or safety purposes, but visual quality for entertainment purposes is paramount. Accurate visual reproduction of fire is prominent in entertainment industries like movie visual effects and video games.

Research techniques related to computer generated fires have been successfully applied in the movie industry. Famous examples include, among others, a planet explosion in Star Trek II, where a particle-based technique by Reeves~\cite{Reeves:1983} was used; Shrek featured a dragon exhaling fire, where parametric curves were used to drive the flames~\cite{Lamorlette:2002}; or the more recent work by Horvath and Geiger~\cite{Horvath:2009} based on 2D screen projections for the film Harry Potter and the Deathly Hallows.
In these and in many other applications, using real flames would have been an expensive and hazardous endeavor.

The computer graphics community has intensively researched the fluid behavior of water and smoke.
While fire can be modeled as a fluid, techniques used for water or smoke cannot be directly applied to flames.
This is due to specific fire proprieties like multi-phase flow, fast chemical reactions and radiative heat transport.
As a result of the complexity and the interdisciplinary nature of the problem, fire simulation is still an open problem in computer graphics.

A great deal of work done in the area has sacrificed complexity for interactiveness, therefore producing simplified models which hope to deceive the observer by exploiting the chaotic behavior present in fire motion.
Nevertheless, physically-based simulations incorporate the intrinsic processes that occur in a combustion scenario in order to be able to produce realistic results.

Current state-of-the-art methods in fluid appearance transfer~\cite{Jamriska:2015,Okabe:2015} overlook the emissive characteristics of fire, be it image based or with full volume reconstruction, the synthesized data fails to illuminate other objects in the scene.
Artists must manually reproduce the aforementioned effects via additional light sources.
Physically-based rendering methods that faithfully recreate the global illumination effects for a flame have been proposed~\cite{Nguyen:2002,Pegoraro:2006}.
However, the input data needed by those techniques are volumetric temperature fields and species concentrations, which can be either synthesized employing complex simulation software~\cite{Uintah} or captured from real flames using equally complex equipment~\cite{Schwarz:1996}.

We propose a new method that can model fire volume with physically valid properties (e.g., temperature fields and species concentrations) from multi-view stereo images.
In our method, the fire shape is represented by a grid volume which is reconstructed using tomographic techniques~\cite{Ihrke:2004}.
The fire appearance is modeled using a physics-based fire renderer~\cite{Pegoraro:2006} which maps the physical properties to the actual appearance. 
The physical properties are estimated by a carefully designed hierarchical optimization referring to the input images, such that the rendered images match the input images in the color space.
We test our method on a variety of inputs and the results show that the generated fire models are not only visually pleasing (see Figure~\ref{fig:teaser}), but can also be used for global fire illumination based visual effects, which is not possible for previous image-based fire modeling tools.

Overall our work makes two major contributions: 1)
 We propose the first image-based fire modeling method that can estimate physical properties instead of color intensities;
2) We demonstrate a novel application of using physically plausible fire generated from images for global illumination based visual effects.
%We propose a new method to estimate physically-plausible volumetric temperature fields and species concentrations from a stereo pair of RGB images of real flames.
%An initial volume is estimated using off-the-shelf tomographic techniques~\cite{Ihrke:2004}.
%Genetic Algorithms are used to find the optimal values in the estimated volume that match the appearance of the input images.
%Our technical contribution lies in the definition of the energy function and novel GA operators.

\section{Related work}

In this section we present an overview of different techniques used for fire rendering and inverse rendering problems.
A more detailed discussion of fire modeling, simulation and rendering is given in Huang et al.~\cite{Huang:2014}.

\subsection{Fire rendering}

Many physically-based methods have been proposed to render participating media realistically. 
Typically, approximate solutions for the Radiative Transport Equation (RTE)~\cite{Howell:2002} are computed.
Rushmeier et al.~\cite{Rushmeier:1995} presented a method to perform accurate ray casting on sparse measured data.
The fire was modeled as a series of stacked cylindrical rings, where each ring has uniform properties.

%The total radiance at each point is integrated using a Monte Carlo method, summing up the measured irradiances at sample locations.
A technique to animate fire with suspended particles was introduced by Feldman et al.~\cite{Feldman:2003}.
Emitted light was computed using Planck's formula of black body radiation, however their RGB mapping requires a manual adjustment using images of real explosions.
%Direct illumination shadows were computed using deep shadow maps~\cite{Lokovic:2000}, while scattering and illumination by other objects in the scene used the technique proposed by~\cite{Jensen:2002}.

Nguyen et al.~\cite{Nguyen:2002} proposed a ray marching technique using black body radiation as well, scattering in the media and the observer's visual adaptation to the fire are modeled.
The visual adaptation method assumes that the hottest part in the single flame maps to white for the observer. 
An extension was presented by Pegoraro and Parker~\cite{Pegoraro:2006}, the authors' model has physically-based absorption, emission and scattering properties.
The spectroscopic characteristics of different fuels are achieved by modeling the transitions of electrons between different energy states in the molecules.
The method allows for non-linear light trajectories  in the medium due to refraction and it includes visual adaptation effects by means of a post-processing mechanism.

\begin{figure*}[t]
\includegraphics[width=\textwidth]{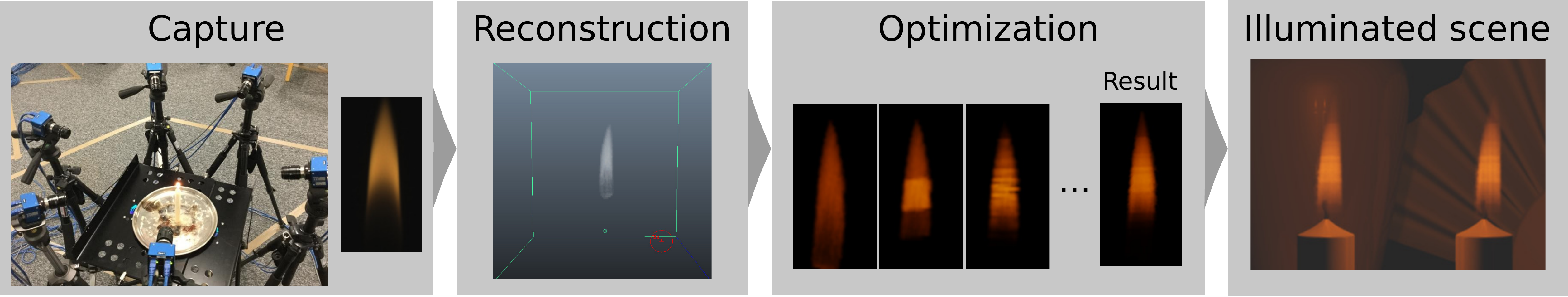}
\caption{Overview of our fire modeling method, from left to right, a multi-view camera setup is used to \emph{capture} real flames, density 3D volumes are computed in the \emph{reconstruction} stage, our \emph{optimization} procedure finds temperatures and fuel density fields that when rendered resemble the captured images, finally the output can be applied for realistic \emph{illumination} of virtual scenes.}
\label{fig:overview}
\end{figure*}

Horvath and Geiger~\cite{Horvath:2009} proposed a rendering method whose main objective was user-friendliness for artists.
The authors perform simple volume rendering on several fixed camera slices to generate an image.
Black body radiation is used to compute light emission; the result is motion-blurred with a filter based on the velocities in the slices, and heat distortion was added as a post-processing filter defined by the user.

Rendering flames at interactive frame rates has also been explored, this techniques inevitably sacrifice quality for performance.
Bridault et al.~\cite{Bridault:2006} used a spectrophotometer to capture photometric distributions of candles.
The intensities are stored on a texture and changes in illumination over time are approximated with an attenuation factor proportional to the size of the flame.
\cite{Zhang:2011} used a plane blending technique were a one-dimensional color texture is used as a transfer function to convert flow attributes to colors and opacities. 

\subsection{Parameter optimization}

Dobashi et al.~\cite{Dobashi:2012} proposed a method to compute eight rendering parameters for clouds shapes using a real cloud photograph.
The authors' technique is limited to simplified scenes with a single light and the cloud, were the camera and the light position are fixed.
Under these restrictions a set of images can be pre-computed to accelerate the optimization.

Klehm et al.~\cite{Klehm:2014} presented a multi-view optimization framework to modify the appearance of volumetric data using input images from several views.
The goal images must perfectly match the shape of the data and the position of the cameras in the scene.
A texture synthesis technique to generate fire animations was introduced by~\cite{Jamriska:2015}.
The method requires a hand made motion field and an alpha mask of the desired result, both of which are used to generate a new sequence using data from an existing video exemplar.

Okabe et al.~\cite{Okabe:2015} presented a technique to reconstruct a volume shape using a sparse set of images.
The texture for the volume is transferred separately using a pyramid-based texture synthesis approach.
Recently a method to generate textures for fluids in animations was proposed by Gagnon et al.~\cite{Gagnon:2016}.
From an initial set of $uv$ coordinates, the new ones are generated by tracking the topological deformations in a number of sample points in the fluid.

Shading parameters for fabrics were estimated using photographs to match appearance and Micro-CT scans to reconstruct fiber-level geometry~\cite{Khungurn:2015}.
The previous methods only consider flames in isolation~\cite{Okabe:2015} or only compute a subset of the parameters, such as the final extinction coefficients at each voxel~\cite{Klehm:2014} or other simpler abstractions~\cite{Dobashi:2012}.
In contrast, our method operates of the realm of physically valid temperatures and fuel densities.
% Check two new citations for Jamriska and one for Okabe
% Add all of them

\section{Methodology}

An overview of the system is shown in Figure~\ref{fig:overview}. 
In the first step we use a multi-view camera setup to record real flames from different viewpoints. To acquire the flames we use a Norpix StreamPix 6 camera system, frame-synchronized at 100 FPS.
For each set of concurrent frames we use 3D tomography techniques~\cite{Ihrke:2004} to obtain a volumetric representation of the flame. 

In the main processing step, the fire appearance is modeled using physical parameters, i.e.~temperature and density fields, and reconstructed based on a carefully designed optimization, which adjusts the parameters such that the appearance of the fire volume coincides with the captured images under corresponding camera views. 
Finally, the reconstructed fire model can be seamlessly integrated in any virtual scene using physically-based fire shaders~\cite{Pegoraro:2006}. 

Due to its physical accuracy, our flame models illuminate the scene naturally and interact with the objects therein without any additional efforts. 
Since the data capture and volumetric reconstruction steps are based on off-the-shelf techniques, we will mainly describe the optimization algorithms for appearance modeling in the following sections.

%An overview of the system is shown in Figure~\ref{fig:overview}.
%The first step is data capture stage, which can consist of recording real flames recorded using a multi-view camera setup.
%A 3D tomographic reconstruction is applied to the input data using Irkhe and Magnor's method~\cite{Ihrke:2004}.
%The reconstructed volume contains an RGB value in each voxel which when projected under each camera view reproduces the input images.
%Our optimization procedure finds plausible temperature and density fields which can replicate the appearance of the input fire.
%These data can then be integrated seamlessly in any virtual scene, using a physically-based shader the flames illuminate the scene naturally and interact with the objects present in their environment.

\subsection{Problem formulation}

For physically-plausible fire appearance modeling, our goal is to find physical parameters of a flame such that when they are used to render an image, the output resembles an input photograph. Given the reconstructed fire volume, the parameters which we are interested in are the camera exposure settings, the flame volumetric temperature,  and fuel density distributions.

\noindent Formally, the above statement can be written as:
\begin{equation}
\label{eq:problem_formualation}
\argmin_{\lbrace	 t, d, s \rbrace} E = \Vert r(t, d, s ) - I_{cam} \Vert, \lbrace t, d \rbrace \in \mathbb{R}^{n_v}, s \in \mathbb{R},
\end{equation}
where $E$ is the energy function to be minimized, $t$ and $d$ are volumes of temperatures and densities with $n_v$ voxels, $I_{cam}$ is the input photograph captured by the camera, $s$ is the exposure of $I_{cam}$, $r: \mathbb{R}^{2 n_v + 1} \mapsto \mathbb{R}^{3 \times n_I}$ is the rendering function that transforms the volumes $t$ and $d$ and the exposure $s$ into an RGB image $I_{cg}$, $n_I$ is the total number of pixels in $I_{cg}$ and $I_{cam}$, and $\Vert \cdot \Vert$ is a similarity metric which compares the source with the target and will be detailed in the next section.

The rendering function $r$ is computed using the Radiate Transfer Equation (RTE)~\cite{Siegel:2002}.
For the absorption and emission coefficients we use the methods described by Pegoraro and Parker~\cite{Pegoraro:2006}.

\subsection{Error function}

In order to estimate the parameters defined in the previous section, we present an error function inspired by Markov Random Fields optimization methods, which define probability estimates for the data and pairwise terms for the variables to be estimated.

The \emph{data term} contains an appearance matching function which ensures that the synthetic image $I_{cg}$ generated by our method matches the camera input image $I_{cam}$.
Histogram distance functions~\cite{Rubner:2000} have been used in previous work~\cite{Dobashi:2012}, however these metrics would fail to transport the complex features present in flames.
Direct pixel to pixel comparison can be applied as the images are aligned. We measure the appearance error as the perceived color difference for each pixel
\begin{equation}
\label{eq:appearance_matching}
E_{am}\left( I_{cg}, I_{cam} \right) = w_{am} \sum_{i = 1}^{n_I} \Vert I_{cg}(i) -  I_{cam}(i) \Vert,
\end{equation}
were $w_{am}$ is a weighting factor, $I_{cg}(i)$ and $I_{cam}(i)$ are the $i^{th}$ pixels values in the CIELab color space~\cite{fairchild2013color} of the synthetic and target images, and the error distance between two pixel colors $C_1$ and $C_2$ is measured as
\begin{equation}
\label{eq:pixel-lab-dist}
\Vert C_1 - C_2 \Vert = \sqrt{ \left( l_1 - l_2 \right)^2 + \left( a_1 - a_2 \right)^2 + \left( b_1 - b_2 \right)^2 }.
\end{equation}

For the \emph{pairwise term} a smoothness estimate is used to avoid unrealistic variations between neighboring voxels, as well as tiling artifacts.
This is justified by the heat diffusion equation: even with the chaotic nature of fire, nearby points in space will have similar temperatures.
Note that the previous statements only hold if the volume resolution is large enough.
The term is computed as
\begin{equation}
\label{eq:gradient_fnc}
E_{sm}\left( v_i, \mathcal{N}_i \right) = w_{sm} \sum_{j\in \mathcal{N}_i} \vert v_j - v_i  \vert,
\end{equation}
where $w_{sm}$ is a weighting factor, $v_i$ is the physical parameter (either a temperature or a density) of the $i$-th voxel, and $\mathcal{N}_i$ is a set with the indices of the 18 closest neighboring voxels of the $i$-th voxel. %The fuel densities also share the same smoothness constraints.
A lower number of neighbors, for instance six would use the immediate adjacent voxel one on each dimension, yet it proved to be insufficient to maintain smoothness in highly disconnected flames, while values larger than 18 become a significant computational burden for the pairwise term.

As several views of the fire volume are available, the method can be easily extended to match the appearance of each of the views.
The total score is computed by integrating the values of each view.
Naturally, the complexity in the evaluation of the data term increases linearly with the number of input images used during the optimization.
For simple and mostly symmetric fires, e.g. candle flames, two views are generally enough to provide good results.
However, when dealing with more complex shapes the number of cameras needed might increase up to six.

Finally, the total score is computed as a linear combination of the previous data and pairwise terms $E = w_{am}E_{am} + w_{sm}E_{sm}$.

\begin{figure}[t]
	\centering
	\includegraphics[width=0.3\linewidth]{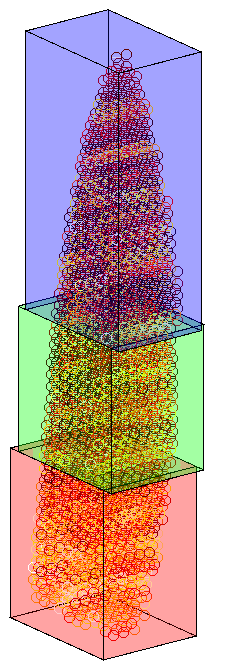} \,
	\includegraphics[width=0.3\linewidth]{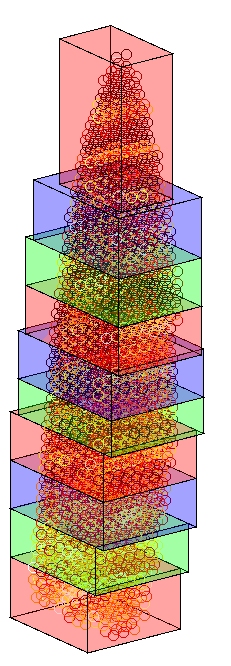} \,
	\includegraphics[width=0.3\linewidth]{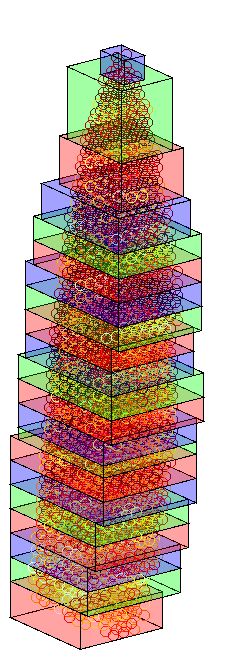}
	\caption{Clustering examples, each cube depicts a different cluster, left to right 3, 10 and 20 divisions.}
	\label{fig:clustering}
\end{figure}

\begin{algorithm}[t]
	\caption{Parameter estimation procedure.}
	\label{tb:icm-pseudocode}
	initialise\_clusters\_and\_other\_variables\;
	\While{not exitConditions()}{
		\For{$i=0$ to $n_{cluster}$ }{
			\For{$j=0$ to $n_{samples}$ }{
				$v_{new} =$ sampleNewValue()\;
				new\_score $=$ dataTerm($v_{new}$) + pairwiseTerm($v_{new}$)\;
				\If{\upshape{new\_score < current\_score(i)}}{
					$(^*x)(i) = v_{new}$\;
					current\_score(i) = new\_score;
				}
			}
		}
		exposure = estimateNewExposure()\;
		updateClustering()\;
		current\_score = updateScore()\;
		total\_score = sum(current\_score)\;
		\eIf(\tcp*[h]{Switch variable}){$x == \&$\upshape{densities}}{
			$x = \&$temperatures;\
		}{
			$x = \&$densities;\
		}
	}
\end{algorithm}

\subsection{Estimation method}

Finding a global solution for the system is intractable due to the non-linearities in the equations that govern flame behavior.
Estimating the Jacobian is computationally expensive.
Therefore we opted for a local gradient sampling strategy based on Coordinate Descent, which allows us to inexpensively evaluate the gradient locally.
For each dimension a simple global evaluation is computed, with the (inaccurate) assumption that there are no dependencies between variables, and we greedily take the value with the lowest error.

The optimization procedure is shown in Algorithm~\ref{tb:icm-pseudocode}, where $\&$ and $(^*)$ denote respectively memory address and pointer dereference operators.
Note that similarly to the expectation maximization (EM) algorithm~\cite{dempster1977maximum}, each collection of variables $t, d$ and $s$ is optimized sequentially, by fixing the value of the remaining ones.

In order to avoid local minima and improve the convergence rate of the algorithm, a clustering approach is used, as shown in Figure~\ref{fig:clustering}.
Initially the number of clusters in the volume is reduced to only two, as we have a sparse representation.
The voxel indices are ordered along the yzx dimensions, with y up and x right.
The first half of the indices will be treated as the first cluster and the rest as the second.
The updateClustering() function increases the number of clusters as the optimization progresses, on each update the resolution is doubled.

The temperature and densities in each voxel, as well as the exposure can take floating point precision values.
From a pure theoretical perspective the number of labels is infinite, and in practice it is too large to evaluate every label on each iteration.
As an approximation we sample around the current point using a Gaussian distribution centered on the current value, with a standard deviation which is inversely proportional to the iteration number.

We model the exposure of the camera by a single floating point value constrained in the range $s \in [0.01,1000]$.
Since modifying the exposure does not modify the voxel neighbor relationships, only the data term is evaluated for each new sample.
If there is a decrease in the total error, the exposure is updated with the new value.

\begin{algorithm}[t]
	\caption{Simplified parameter estimation procedure.}
	\label{tb:icm-common-d-pseudocode}
	initialise\_clusters\_and\_other\_variables\;
	\While{not exitConditions()}{
		\For{$i=0$ to $n_{cluster}$ }{
			\For{$j=0$ to $n_{samples}$ }{
				$v_{new} =$ sampleNewValue()\;
				new\_score $=$ dataTerm($v_{new}$) + pairwiseTerm($v_{new}$)\;
				\If{\upshape{new\_score < current\_score(i)}}{
					temperatures$(i) = v_{new}$\;
					current\_score(i) = new\_score;
				}
			}
		}
		density\_factor = estimateNewDensityFactor()\;
		exposure = estimateNewExposure()\;
		updateClustering()\;
		current\_score = updateScore()\;
		total\_score = sum(current\_score)\;
	}
\end{algorithm}

\begin{figure*}[t!]
        \centering
        \begin{subfigure}[t]{0.23\linewidth}
                \includegraphics[height=3.25cm]{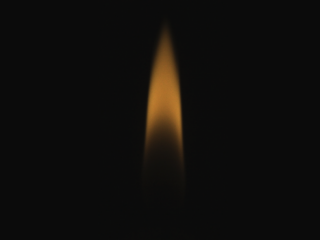}
                \caption{Candle fire, front view.}
        \end{subfigure}
        \,
        \begin{subfigure}[t]{0.23\linewidth}
                \includegraphics[height=3.25cm]{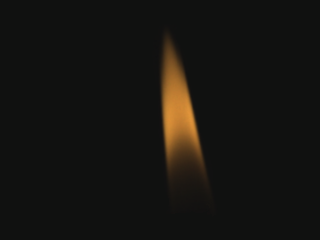}
                \caption{Candle fire, side view.}
        \end{subfigure}
        \,
        \begin{subfigure}[t]{0.23\linewidth}
                \includegraphics[height=3.25cm]{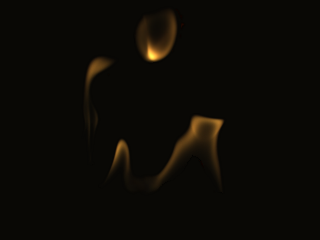}
                \caption{Fire licks, front view.}
        \end{subfigure}
        \,
        \begin{subfigure}[t]{0.23\linewidth}
                \includegraphics[height=3.25cm]{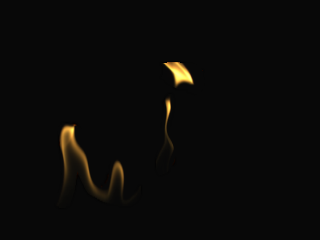}
                \caption{Fire licks, side view.}
        \end{subfigure}
        \caption{Goal images for our data, (a) and (b) front and side view of a candle flame, (c) and (d) front and side view of several fire licks.}
        \label{fig:goal-images}
\end{figure*}

In the initialization step, each voxel is assigned a temperature and density randomly sampled with uniform probability from a range of maximum and minimum values which are given by the physical nature of the data~\cite{Howell:2002}, $t \in [300, 2300]$ degrees Kelvin and $d \in [0.01 \times 10^{27}, 500 \times 10^{27}]$ number of particles per cubic meter.
The initial exposure is sampled using a logarithmic scale from its own bounded range and the exposure value with the lowest error is chosen.
Once the initial value for the exposure has been estimated, the mean value for $t$ and $d$ of the voxels that compose a cluster is assigned to the cluster itself.
The standard deviation of all the Gaussian distributions that are sampled during the optimization is progressively decreased as the number of iterations increases.

In practice a simplified optimization method can be used instead of the one shown in Algorithm~\ref{tb:icm-pseudocode}.
The RGB values can be a good approximation to the real fuel density field.
We take the red channel of each voxel from the volumetric reconstruction stage and we normalize all the values.
During the optimization a single density scale factor is estimated instead of the full range of possible densities per voxel in a step analogous to the exposure estimation.
This variation of the optimization procedure is shown in Algorithm~\ref{tb:icm-common-d-pseudocode}.
With this approximation the dimensionality of the parameters to be estimated is reduced from $2 n_v +1$ to $n_v + 2$, i.e. the amount of data renderings is reduced by half.
In our experiments this simplification was able to yield compelling results.

\section{Results and discussion} 
\label{sec:results-and-discussion}

For the results shown in this section we used a server with 24 Intel Xeon E5-2620v2 2.10GHz processors and 62GB of RAM.
The input parameters for all the figures in the paper were $w_{am} = 1$ and $w_{sm} = 10$.
The volume data for the flames fuel density and temperature distribution have a resolution of $128 \times 128 \times 128$ voxels.

The images were assembled in Maya 2015 and rendered using a custom Mental Ray CPU shader which implements ray marching for fire rendering based on the work presented by Pegoraro and Parker~\cite{Pegoraro:2006}.

The evaluation times range from two to twelve hours using an unoptimized Matlab CPU implementation, where the main bulk of the computation time goes into the fitness evaluation, i.e. rendering one image with a set of input parameters.

The images used during the optimization have a resolution of $320 \times 240$ pixels, while the results which show the output flames in more complex scenes were rendered at a higher quality, $960 \times 540$. 

For the figures shown in the paper, the goal images were taken from the frontal and side (90 degrees) cameras, as shown in Figure~\ref{fig:goal-images}.

The optimization time is proportional to the image and volume sizes, as well as the sampling rates for the optimization and rendering parameters.
Unless otherwise stated all the results were generated with Algorithm~\ref{tb:icm-common-d-pseudocode}.

The voxels in the volume are flattened into an array and are ordered along the yzx dimensions, with y up and x right.
For a $128 \times 128 \times 128$ voxel space, each temperature and density volumes generated after applying the aforementioned operation is $2097152$ dimensional, which can be challenging to optimize.
To address this issue, we use a sparse representation for the volume data.
After the reconstruction step any RGB voxel value that is below a certain threshold is considered to be empty, which reduces the dimensionality of the data during the search.

\begin{figure}[t]
        \centering
        \begin{subfigure}[t]{0.24\textwidth}
                \includegraphics[width=\textwidth]{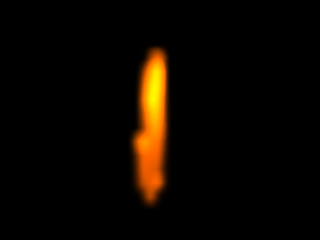}
                \caption{Goal image.}
        \end{subfigure}%
        ~
        \begin{subfigure}[t]{0.24\textwidth}
                \includegraphics[width=\textwidth]{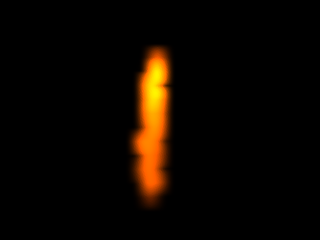}
                \caption{Result image.}
        \end{subfigure}
        \caption{Fire modeling with our method using synthetic data, (a) the goal image rendered with synthetic data and (b) the optimization result.}
        \label{fig:ground-truth-results}
\end{figure}

To be able to validate our method, we initially optimized the parameters with respect to goal images rendered from synthetic data.
The RGB volume extracted from the multi-view 3D reconstruction algorithm from a video of a real flame is used as initialization.
The density field is the R component in each voxel scaled to a plausible density, the same procedure is applied to the temperature using the maximum of  each channel.
The goal image is the result of rendering the previous data with a manually tuned exposure.
Figure~\ref{fig:ground-truth-results} shows the result of the optimization via Algorithm~\ref{tb:icm-pseudocode} with a volume resolution of $32 \times 32 \times 32$ voxels.

Given the non-linear interaction between the input parameters in the RTE and the ambiguity introduced by 2D projection for image rendering, it can be assumed that the optimization space has many local minima.
As a result different initialization parameters may lead to different outputs.
However, our objective is not to accurately estimate the temperature and density fields, but to produce plausible estimates which will faithfully reproduce the input images.
And as such, those variations are satisfactory for our purposes.

\begin{figure}[t]
\centering
\includegraphics[width=0.8\linewidth]{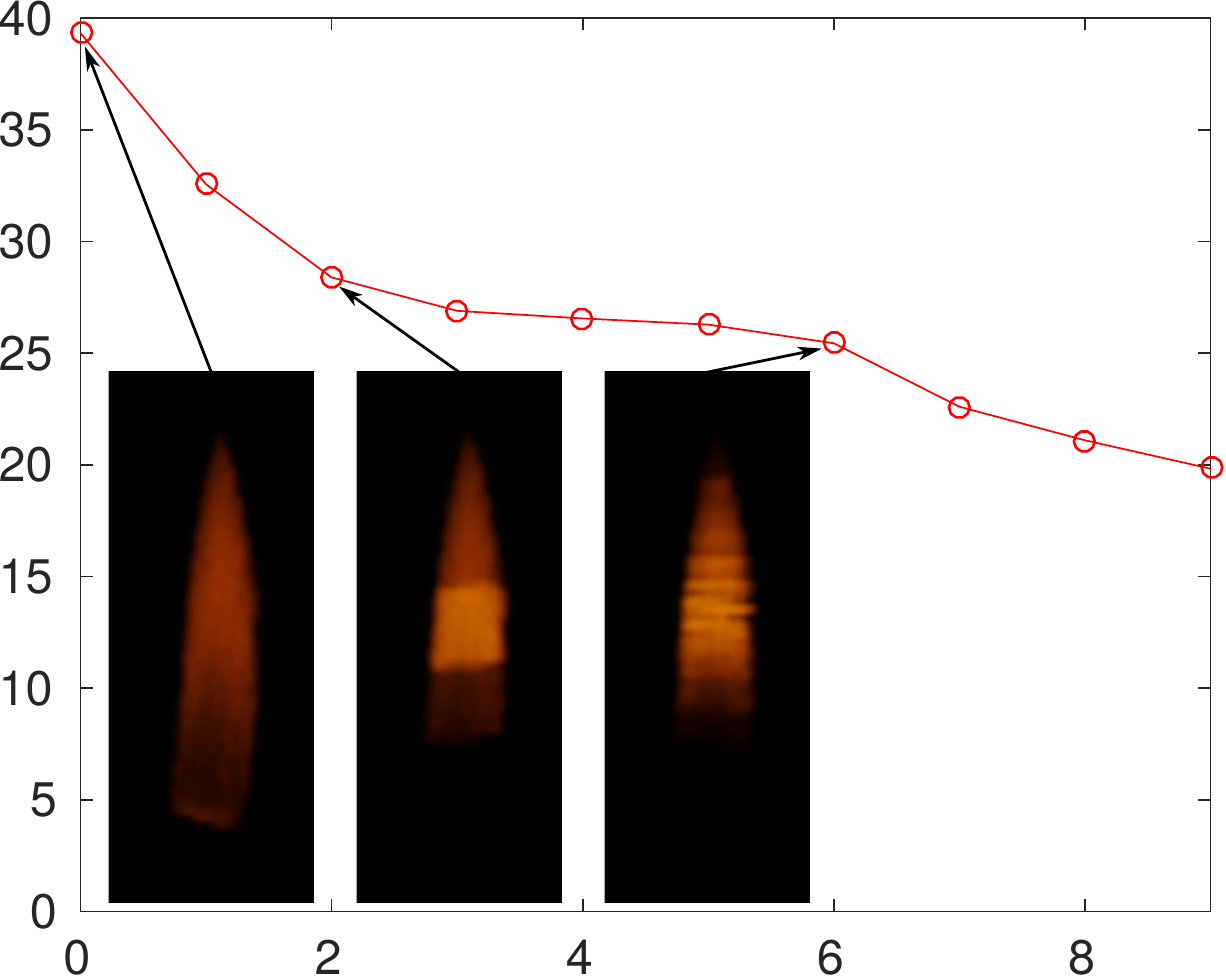}
\caption{Transition of the error function during the optimization.}
\label{fig:objective-function}
\end{figure}

The evolution of the temperature values during the optimization with their corresponding fitness value is shown in Figure~\ref{fig:objective-function}.
The figure shows how the objective function decreases rapidly in the first few iterations and stabilizes as the optimization progresses.
Note how our clustering approach first finds the right color for large regions, which are progressively refined as the resolution increases. 

Analyzing the spatial distribution of the output volumes after each iteration gives some insight into how the Coordinate Descent method explores the error function space.
This test reveals how overall variations in temperatures and densities map to changes in the error function, which in turn is a measure of color variation.

\begin{figure}[t]
\includegraphics[width=0.99\linewidth]{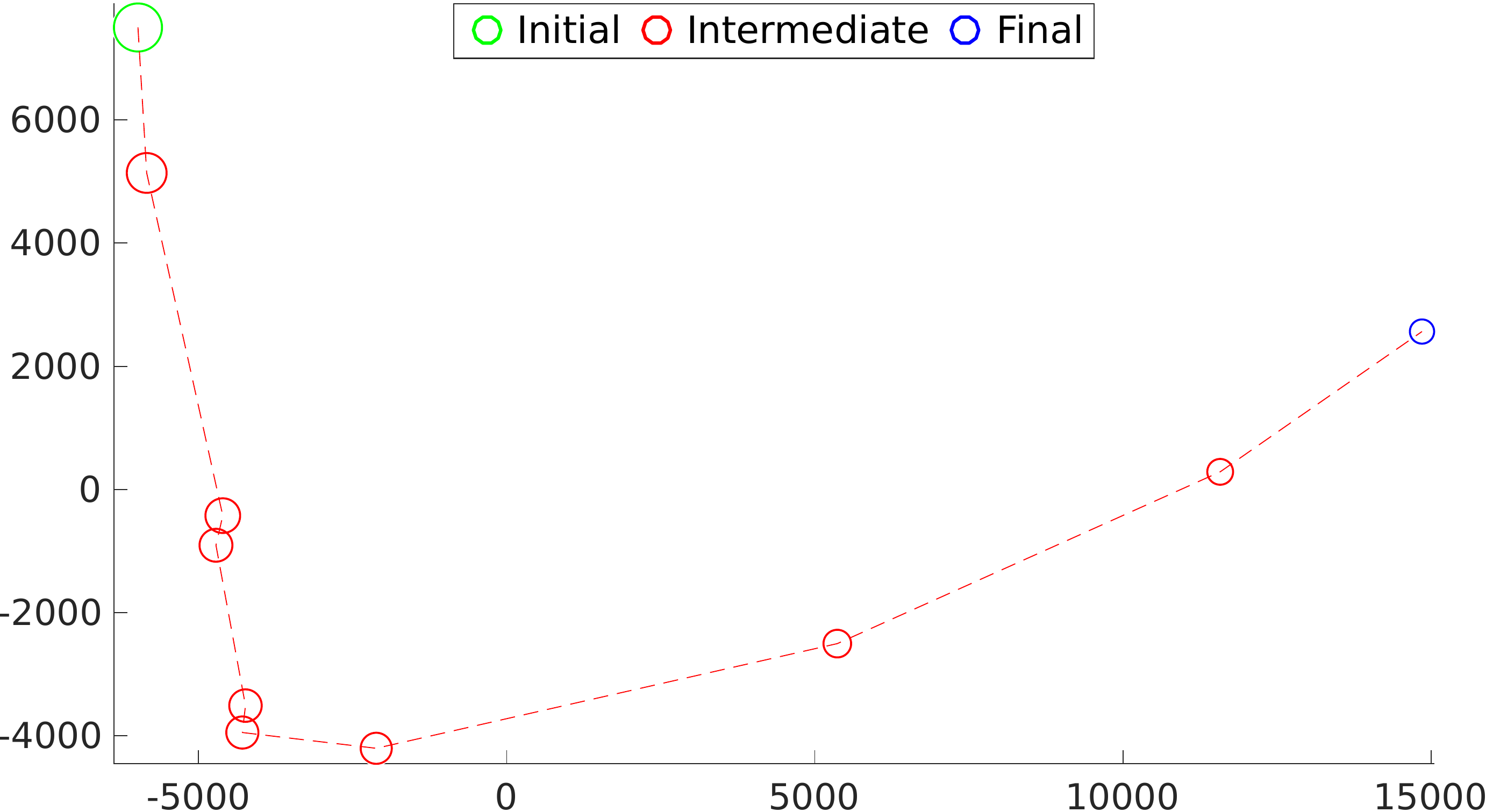}
\caption{Temperature fields during the optimization projected on a 2D plane, x and y axes are euclidean distance in temperature space and the radius of the circles are proportional to the value of the error function.}
\label{fig:mds1}
\end{figure}

Each flame is $2n_v+1$-dimensional, to simplify the interpretation of the data, only the $n_v$-dimensional temperature fields are depicted in Figure~\ref{fig:mds1}.
Each circle represents a flame temperature in a given iteration, MultiDimensional Scaling~\cite{Seber:1984} is applied to project the data into a 2D space, where the axes are in some arbitrary units which preserves the Euclidean distances as measured in the parameters space.
The radius of the circles are proportional to the score of the error function, and the lines connect adjacent iterations.
It can be seen clearly, especially when making a comparison to Figure~\ref{fig:objective-function}, that a large variation in the temperature field does not have to necessarily imply large variations in the error function.

\begin{figure}[t]
        \centering
        \begin{subfigure}[t]{0.24\textwidth}
                \includegraphics[width=\textwidth]{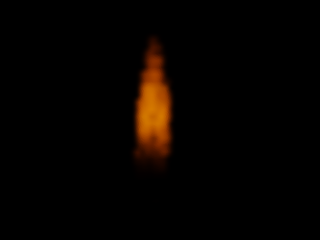}
                \caption{Without smoothness.}
                \label{fig:optimized-Cam1-without-neigh}
        \end{subfigure}%
        ~ 
        \begin{subfigure}[t]{0.24\textwidth}
                \includegraphics[width=\textwidth]{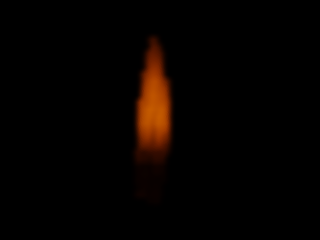}
                \caption{With smoothness.}
                \label{fig:optimized-Cam1-with-neigh}
        \end{subfigure}
        \caption{Effects of the smoothness term, (a) output flame without the smoothness term and (b) result with the smoothness term.} 
        \label{fig:smoothness-effects}
\end{figure}

To demonstrate visually the effects of each term in the error function, we optimized only with the data term, the result is shown in Figure~\ref{fig:optimized-Cam1-without-neigh}.
Note how the flame has drastic temperature changes between neighboring voxels and the shape has little resemblance to the original fire.
The same optimization adding the pairwise smoothness term to the objective function is shown in Figure~\ref{fig:optimized-Cam1-with-neigh}.
The shape now better resembles the original, and the overall color remains mostly the same.

\begin{figure}[t]
        \centering
        \begin{subfigure}[t]{0.24\textwidth}
                \includegraphics[width=\textwidth]{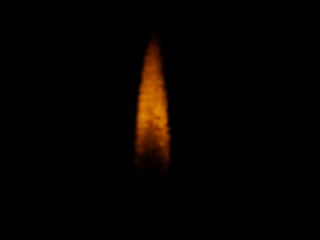}
                \caption{Single goal, optimized view.}
        \end{subfigure}%
        ~ 
        \begin{subfigure}[t]{0.24\textwidth}
                \includegraphics[width=\textwidth]{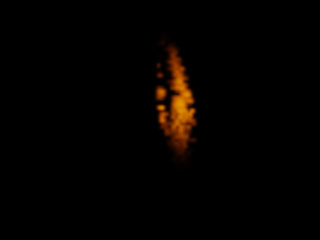}
                \caption{Single goal, novel view.}
                \vspace*{2mm}
        \end{subfigure}
        \\
        \begin{subfigure}[t]{0.24\textwidth}
                \includegraphics[width=\textwidth]{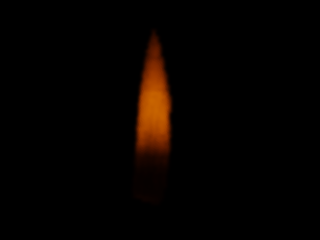}
                \caption{Two goal, optimized view.}
        \end{subfigure}%
        ~ 
        \begin{subfigure}[t]{0.24\textwidth}
                \includegraphics[width=\textwidth]{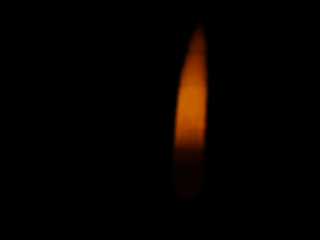}
                \caption{Two goal, novel view.}
        \end{subfigure}
        \caption{Comparison of using single goal, (a) optimized view and (b) novel view, vs multi-view optimization with (c) optimized view and (d) the novel view.} 
        \label{fig:single-multi-view}
\end{figure}

The number of views used as goal images also plays an important role in the optimization procedure.
The output results for a simple candle flame using a single frontal view and, a pair of frontal and side views is shown in Figure~\ref{fig:single-multi-view}.
Note that in the single view setting there are more degrees of freedom, which allows the algorithm to better match the shape and color of the input image.
However, severe artifacts including discontinuities can be observed from other viewing angles.
Using two goal images for the modeling is enough in this scenario to produce satisfactory renders from novel views.

\begin{figure}[t]
	\centering
	\includegraphics[width=0.24\linewidth, trim={3cm 0.5cm 3.5cm 0.5cm}, clip]{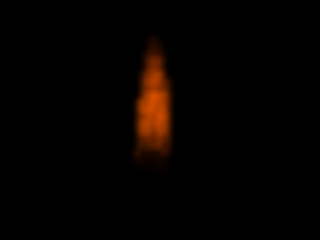}
	\includegraphics[width=0.24\linewidth, trim={3cm 0.5cm 3.5cm 0.5cm}, clip]{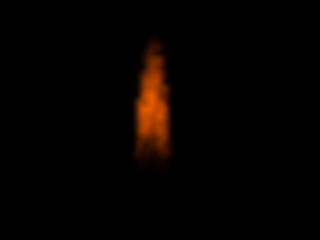}
	\includegraphics[width=0.24\linewidth, trim={3cm 0.5cm 3.5cm 0.5cm}, clip]{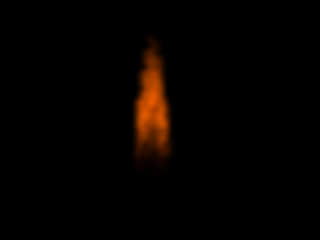}
	\includegraphics[width=0.24\linewidth, trim={3cm 0.5cm 3.5cm 0.5cm}, clip]{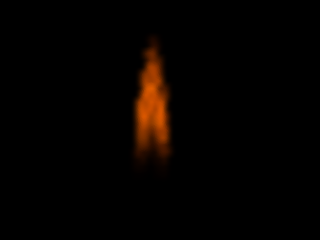}
	\includegraphics[width=0.24\linewidth, trim={3cm 0.5cm 3.5cm 0.5cm}, clip]{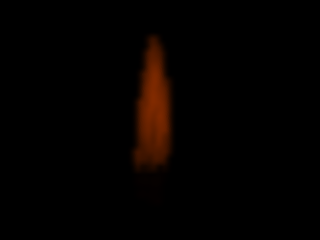}
	\includegraphics[width=0.24\linewidth, trim={3cm 0.5cm 3.5cm 0.5cm}, clip]{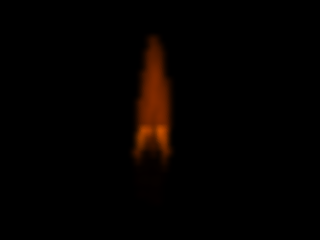}
	\includegraphics[width=0.24\linewidth, trim={3cm 0.5cm 3.5cm 0.5cm}, clip]{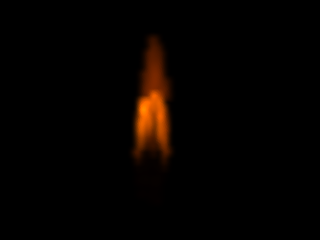}
	\includegraphics[width=0.24\linewidth, trim={3cm 0.5cm 3.5cm 0.5cm}, clip]{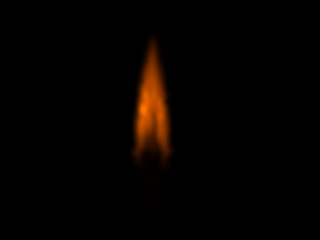}
	\caption{The effects of the clustering on the optimization, top with clustering, bottom without, left to right, after 30 minutes, 1 hour, 2 hours and 3 hours of processing. It is evident how much clustering improves the efficiency of the computation.}
	\label{fig:no_clustering}
\end{figure}

The clustering approach plays a crucial part in the optimization procedure.
Otherwise the system is more likely to fall in worse local minima, as it loses the power to induce large changes in the rendered images.
A comparison of an optimization with and without this feature is shown in Figure~\ref{fig:no_clustering}.
The clustered version reaches lower error and more visually pleasing results with significantly fewer function evaluations.
For this particular experiment the temperature and density volumes were downsampled to a resolution of $64 \times 64 \times 64$ voxels in order to better show the difference between the images.
 
\begin{figure*}[p]
\centering
    \begin{subfigure}[t]{0.7\textwidth}
    		\centering
		\includegraphics[width=\textwidth]{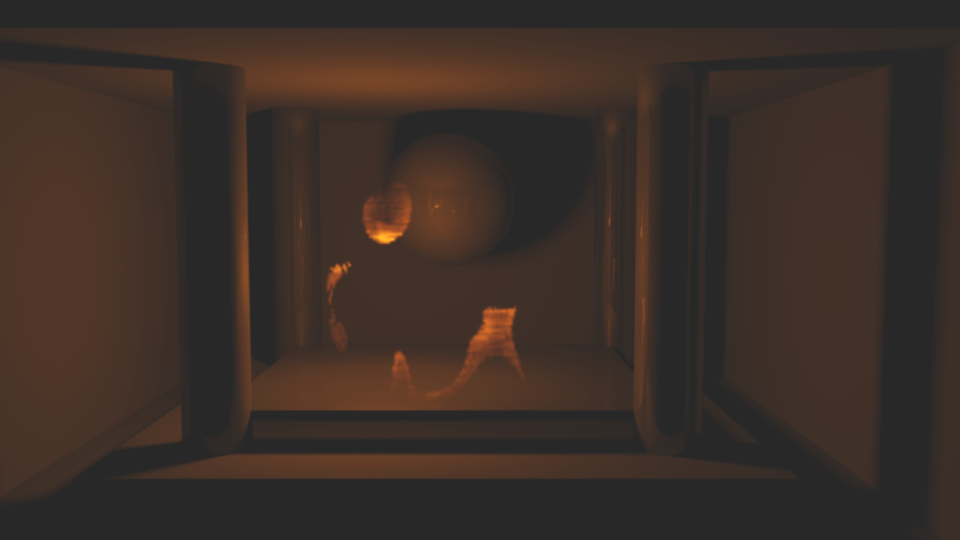}
		\caption{Scene rendered using small spherical light for illumination.}
		\vspace*{2mm}
	\end{subfigure}\\
	\begin{subfigure}[t]{0.48\textwidth}
		\centering
		\includegraphics[height=4.4cm, trim={13cm 9cm 12cm 3cm},clip]{images/teaser1}
		\caption{Shadow detail with our method, patch extracted from Fig~\ref{fig:teaser} left.}
	\end{subfigure} ~
	\begin{subfigure}[t]{0.48\textwidth}
		\centering
		\includegraphics[height=4.4cm, trim={13cm 9cm 12cm 3cm},clip]{images/four-columns-point-light}
		\caption{Shadow detail with spherical light.}
	\end{subfigure}
\caption{Comparison between physically-based illumination accurately sampled in the flame Fig.~\ref{fig:teaser} and an approximation where a spherical shape light is used, (a) the scene with the sphere light, (b) detailed section from (a) and (c) detailed area from~Fig.~\ref{fig:teaser}. Overexposed results, see Fig.~\ref{fig:image-fake-ilum-original} for the unaltered images.}
\label{fig:image-fake-ilum} 
\end{figure*}

To further demonstrate the advantages of our method, we compare the  traditional approach for flame modeling in Maya with ours.
Figure~\ref{fig:teaser} was rendered in Maya with flame emissivity manually disabled, i.e. it would not illuminate other objects in the scene, and a spherical light with the output color of the voxel with the highest temperature was placed in the center of the flame volume.
The result is shown in Figure~\ref{fig:image-fake-ilum}, note how the shadows are not as soft and the areas with more illumination do not correspond to the flame position.
These defects become more apparent as the flame complexity increases, and in simple fires, e.g. candle flames, they are hardly noticeable.

\begin{figure*}[p]
	\centering
	\includegraphics[width=0.7\linewidth]{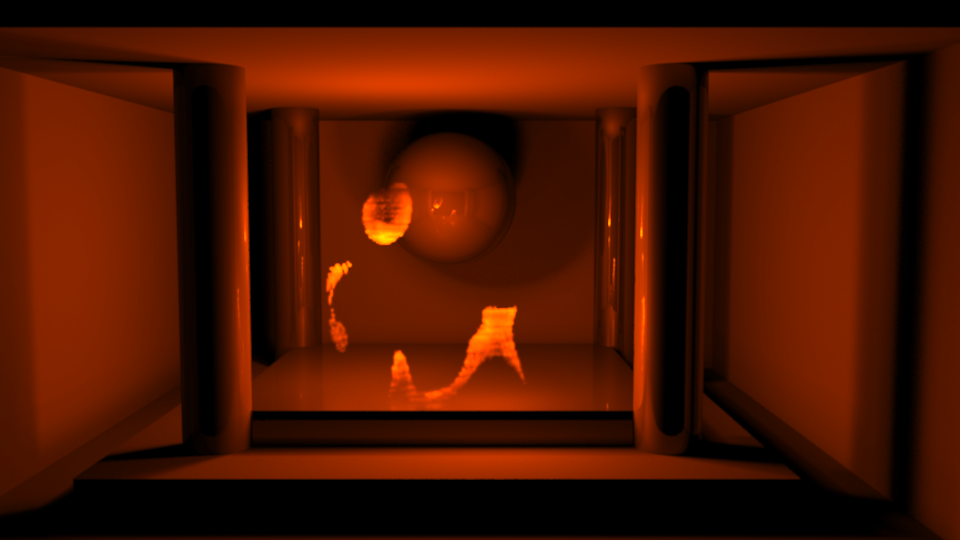}
	\caption{Modeling a different eye adaption level to the flame stimulus using the tone mapping technique of Reinhard et al.\cite{Reinhard:2002} .}
\label{fig:light_adaptation} 
\end{figure*}

Other benefits of our system include the ability to model the human eye visual adaptation to different illumination stimulus via HDR to LDR conversion techniques~\cite{Banterle:2011}.
An example using Reinhard et al.\cite{Reinhard:2002} method is shown in Figure~\ref{fig:light_adaptation}, the effect is comparable to increasing the exposure in the camera.

\section{Conclusion and future work}

We have presented a novel method to model flames from multi-view images. 
Our system can generate physically-based models, it estimates plausible camera exposure, 3D volumetric temperature and density fields from input images.
We allow artists to seamlessly insert flames as seen in input videos into their virtual scenes, and those flames realistically illuminate the scene without the need to artificially add any additional light sources. 
Our work is freely available for usage in further research or industrial projects. 
The source code and data are available at \sourceCodeURL.

There are different opportunities for future work.
A first improvement might be to use the results of the computation at one timestamp to initialize the computation for the following frame, in order to have smoother animations and improved convergence rates. 
One way to achieve this might be via extrapolation of the voxel values between frames. 

Another possible improvement could be to reduce the number of cameras needed to perform the initial volume reconstruction, in order to increase applicability of our system. 
This could be done by integrating techniques such as Okabe et al.~\cite{Okabe:2015} into our framework.

An interesting possible application of our system is to use the output model to initialize conditions and parameters for an existing fire simulation software (e.g.~\cite{Uintah}).
This could allow for mixed techniques that generate new flames from existing video examples.

Lastly, while our system is able to provide visually pleasing and realistic results, improvements can be pursued toward more efficient computation times. Areas of possible improvement include optimized GPU implementations and improved clustering with, for example PCA to compute axes of major change or with k-nearest neighbors.
In the future, this would possibly allow for real-time model computation.

%Although we can constrain the initialization of our method using the constrains from physically realistic flames, the optimization itself is not guaranteed to converge to the global minima, and it can oscillate between states.
%However, in practice we found that the method always converges to a sensible solution.

\bibliography{bibliography}
\bibliographystyle{ieee}

\begin{figure*}[p]	
   \centering
   \includegraphics[width=0.48\textwidth]{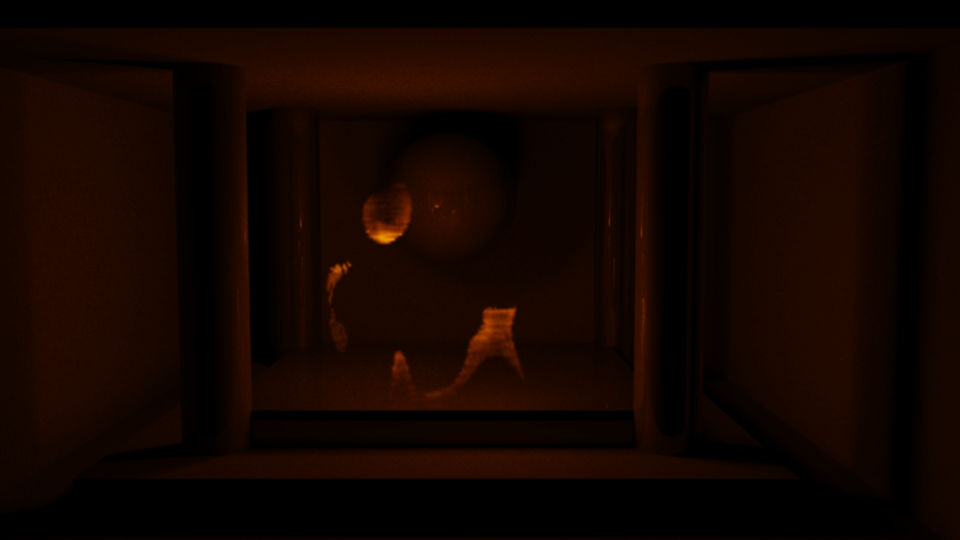}
   \includegraphics[width=0.452\textwidth]{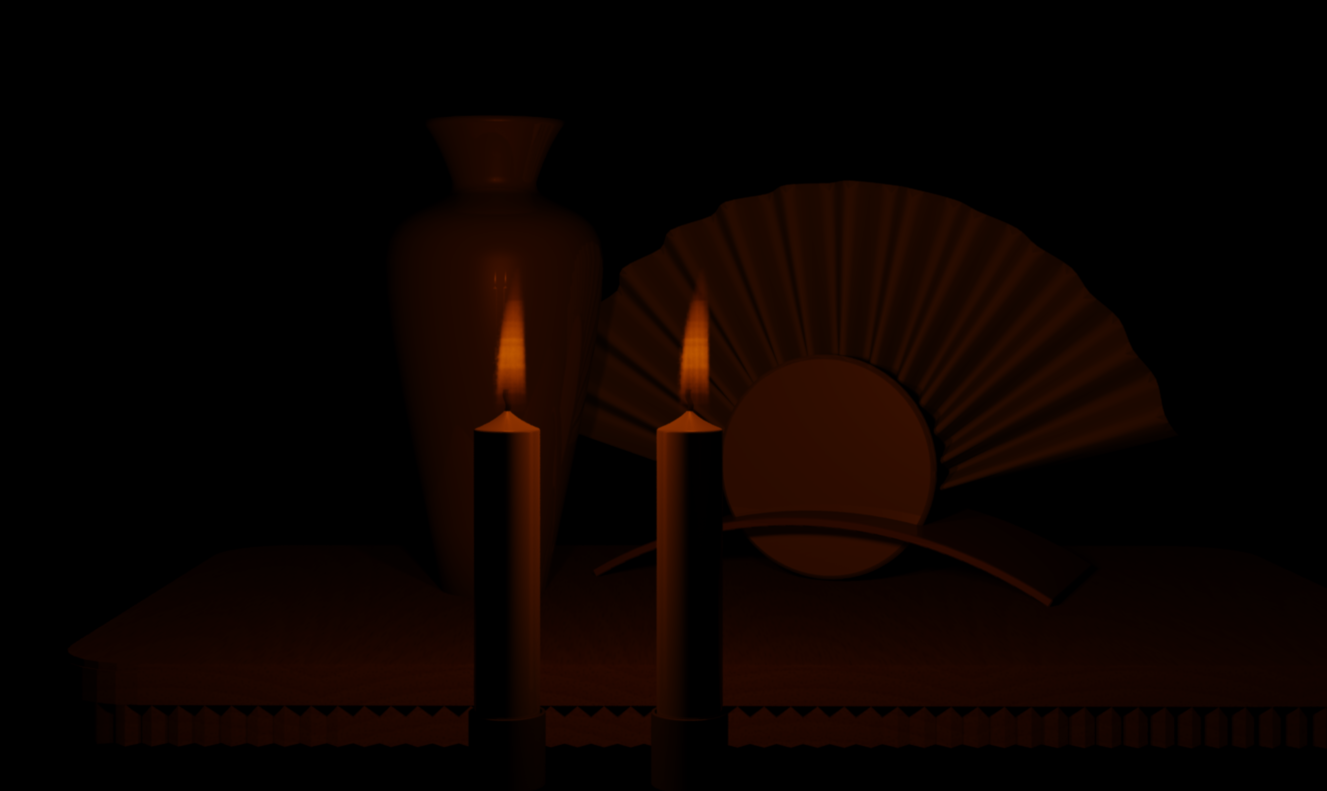}
   \caption{Same output as Figure~\ref{fig:teaser} without overexposure.}
   \label{fig:teaser-original}
\end{figure*}

\begin{figure*}[p]
\centering
    \begin{subfigure}[t]{0.7\textwidth}
    		\centering
		\includegraphics[width=\textwidth]{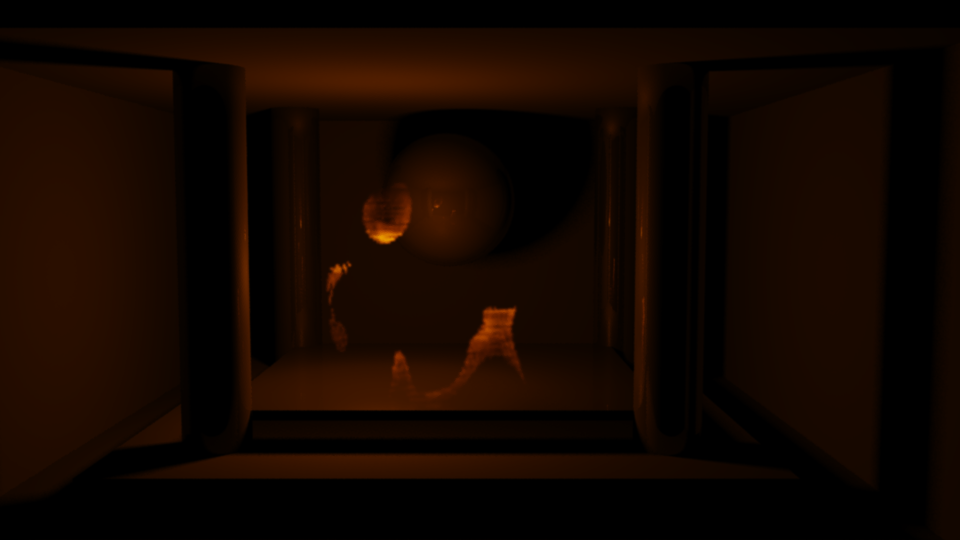}
		\caption{Scene rendered using small spherical light for illumination.}
		\vspace*{2mm}
	\end{subfigure}
	\begin{subfigure}[t]{0.48\textwidth}
		\centering
		\includegraphics[height=4.4cm, trim={13cm 9cm 12cm 3cm},clip]{images/teaser1-original}
		\caption{Shadow detail with our method.}
	\end{subfigure} ~
	\begin{subfigure}[t]{0.48\textwidth}
		\centering
		\includegraphics[height=4.4cm, trim={13cm 9cm 12cm 3cm},clip]{images/four-columns-point-light-original}
		\caption{Shadow detail with spherical light.}
	\end{subfigure}
	\caption{Same output as Figure~\ref{fig:image-fake-ilum} without overexposure.}
\label{fig:image-fake-ilum-original} 
\end{figure*}

\end{document}